%% file: main_DAC.tex
\documentclass[conference]{IEEEtran}
\IEEEoverridecommandlockouts

\usepackage{cite}
\usepackage{amsmath,amssymb,amsfonts}
\usepackage{algorithmic}
\usepackage{graphicx}
\usepackage{textcomp}
\usepackage{xcolor}
\def\BibTeX{{\rm B\kern-.05em{\sc i\kern-.025em b}\kern-.08em
    T\kern-.1667em\lower.7ex\hbox{E}\kern-.125emX}}
\usepackage[switch,mathlines]{lineno}
\usepackage{amsmath,amsfonts}
\usepackage[caption=false,font=normalsize,labelfont=sf,textfont=sf]{subfig}
\usepackage{verbatim}
\usepackage{graphicx}

\usepackage{booktabs}
\usepackage{multirow}
\usepackage[hidelinks]{hyperref}
\usepackage{xcolor}
\usepackage{makecell}

\begin{document}

\title{GreenDynaMix: Thermal-Aware Dynamic Mixed-Precision MoE LLM Inference Systems\vspace{-12.5pt}}

\author{\IEEEauthorblockN{Han-Yi Lin\textsuperscript{1}, Kieran Griperay\textsuperscript{1}, Qiang Zhou\textsuperscript{1}, Xiangyu Zhou\textsuperscript{2}, Nate Bonacci\textsuperscript{1}, Zinachidi Achikeh\textsuperscript{3}\\Li He\textsuperscript{4}, Shaofei Qi\textsuperscript{4}, Hongqiang Wang\textsuperscript{4}, Jeff Zhang\textsuperscript{2}, Tianlong Chen\textsuperscript{5}, Pi-Cheng Hsiu\textsuperscript{6} and Jingtong Hu\textsuperscript{1}}
\IEEEauthorblockA{\textsuperscript{1}\textit{University of Pittsburgh}, \textsuperscript{2}\textit{University of Arizona}, \textsuperscript{3}\textit{Carnegie Mellon University}, \textsuperscript{4}\textit{Qualcomm}\\\textsuperscript{5}\textit{The University of North Carolina at Chapel Hill}, \textsuperscript{6}\textit{Academia Sinica}}}
\IEEEaftertitletext{\vspace{-1\baselineskip}}
\def\IEEEtitletopspace{0pt}
\makeatletter\def\@IEEEMINtitlevspace{5pt}\makeatother
\maketitle

\begin{abstract}
Sustained Large Language Model (LLM) inference generates significant heat that exacerbates silicon aging and reduces GPU reliability. Because GPU temperature fluctuates continuously with workload intensity and ambient conditions, keeping it in a safe range requires real-time frequency scaling, which inevitably comes at the cost of either throughput or accuracy. To address this tightly coupled challenge, we propose \textsc{GreenDynaMix}, the first runtime system to jointly integrate a DVFS-based thermal manager with a dynamic precision-tuning throughput compensator. A sensitivity-aware allocator leverages intrinsic importance signals from the Mixture-of-Experts (MoE) router to assign higher precision to critical experts, preserving accuracy. Experiments on server, desktop, and embedded platforms demonstrate that \textsc{GreenDynaMix} reduces the temperature-induced aging rate of GPUs during LLM inference by 25--33\%. Without offline profiling or model retraining, it recovers throughput at a perplexity cost of 0.9--1.6\% with 8-bit/4-bit mixed precision, rising to 9.4\% where memory forces 4-bit/2-bit.
\end{abstract}


\input{Sections/Introduction}
\input{Sections/Background}

\input{Sections/Observation}

\input{Sections/Approach}

\input{Sections/Evaluation}
\input{Sections/Conclusion}
\clearpage
\bibliographystyle{IEEEtranS}
\bibliography{sample-base,example_paper}

\end{document}

%% file: Sections/Introduction.tex
\section{Introduction}
\label{sec1:Introduction}
Large Language Models (LLMs) are now deployed across datacenters~\cite{zhao2024alise}, local desktops, and embedded and edge platforms~\cite{yu2024edge,wang2025empowering}. Because LLM decoding is a long-running, token-by-token process with continuous back-to-back kernel execution, GPU inference on all of them produces substantial, continuous heat, while cooling capability shrinks from datacenters to space-constrained desktops and passively cooled embedded devices~\cite{prakash2016improving}. Sustained high temperature not only triggers thermal throttling but also accelerates device aging through mechanisms such as electromigration and dielectric breakdown~\cite{kong2012recent,mishra2016predicting,kumar2006analytical}. {\color{black}This dependence has been confirmed at scale: a survival analysis of some 18,000 GPUs on the Titan supercomputer found GPU failure hazard rates correlating strongly with the cooling architecture and operating temperature~\cite{ostrouchov2020gpu}.} Effective thermal management is thus a fundamental requirement for reliable and sustainable LLM inference.

Most prior work on efficient LLM systems~\cite{kim2025efficient} accelerates inference through quantization, pruning, distillation~\cite{zhou2026baton}, and sparse conditional computation such as Mixture-of-Experts (MoE){\color{black}, which has become a mainstream architecture for scaling modern LLMs~\cite{cai2025survey}}. MoE activates only a subset of experts per token, allowing the model to retain a large capacity while computing only a fraction of its weights. Although these techniques finish jobs sooner, they do not mitigate the temperature ramp rate or suppress temperature fluctuations, and as inference demand grows, devices still operate at high temperatures for extended periods. Built-in thermal management, in turn, prioritizes performance over long-term reliability: the NVIDIA RTX~4090 governor uses $85\,^{\circ}\mathrm{C}$ as its default trigger~\cite{lu2025thermal}, without explicitly accounting for cumulative silicon aging, and exposes only a $65$--$88\,^{\circ}\mathrm{C}$ controllable range~\cite{nvidia_gtt_forum}. For LLMs, recent work schedules requests across servers to prevent heat buildup~\cite{stojkovic2025tapas,lu2025thermal}. {\color{black}A second, larger line instead cuts serving energy by operating at the minimum GPU frequency that still satisfies request-level latency service-level objectives, via iteration-level Dynamic Voltage and Frequency Scaling (DVFS), predictive throttling, or cluster-scale reconfiguration of instances, parallelism, and frequency~\cite{kakolyris2024slo,stojkovic2025dynamollm,kakolyris2025throttll,liu2026greenllm}.} However, these schemes are tailored to datacenters and not applicable to local desktop or mobile devices, {\color{black}do not regulate the device to a target temperature,} and do not co-optimize LLM throughput and accuracy, {\color{black}nor are they tailored to MoE.}

Sustainable LLM inference is fundamentally a coupled control problem across temperature, throughput, and accuracy, rather than a single-objective optimization problem. During sustained LLM decoding, GPU temperature evolves continuously with workload intensity and ambient conditions, so thermal control must adjust the operating frequency at runtime, and each adjustment changes the available compute capacity and immediately affects token throughput. Because the frequency needed to hold the temperature is not fixed but varies over time, sustaining throughput requires an equally dynamic compensation mechanism. Such compensation must modulate per-token computation, for example, through dynamic mixed-precision execution, which can in turn compromise accuracy. These challenges therefore cannot be addressed independently: without coordinated runtime support, the system must sacrifice throughput or accuracy to stay thermally safe, or abandon thermal control and accelerate device aging.

To address this dilemma, we propose \textsc{GreenDynaMix}, a thermal-aware runtime for MoE LLMs that integrates three cooperating modules. (1) The \emph{Thermal Manager} continuously adjusts GPU frequency via DVFS to keep the device temperature within a user-specified target. (2) Because each frequency adjustment alters the inference speed, the \emph{Throughput Compensator} dynamically modulates the ratio of high- and low-precision experts during decoding to recover throughput under thermal constraints. (3) Because lowering precision risks accuracy degradation, the \emph{Sensitivity-Aware Allocator} {\color{black}leverages the MoE router's gating weights as an intrinsic importance indicator,} assigning limited high precision to the most important activated experts while leaving the remaining experts in lower precision. Together, these modules form a unified runtime that maintains thermal safety while jointly preserving throughput and accuracy.

Our main contributions are summarized as follows:
\begin{itemize}
    \item \textsc{GreenDynaMix} is the first unified runtime that dynamically co-manages temperature, throughput, and accuracy for MoE LLM inference systems.
    \item {\color{black}We propose a routing-weight-guided, per-token expert-wise mixed-precision scheme that allocates a limited high-precision budget, better preserving accuracy than prior mixed-precision methods at equal budget.}
    \item We implement and evaluate \textsc{GreenDynaMix} on server, desktop, and embedded GPUs, where it significantly reduces the temperature-induced aging rate during LLM inference by 25--33\%, while maintaining throughput and incurring a slight accuracy drop (0.9--1.6\% in perplexity) under 8-bit/4-bit mixed precision.
\end{itemize}

%% file: Sections/Background.tex
\section{Background}
\label{sec2:Background}
\subsection{Thermal Behavior of LLM Inference}
LLM decoding issues GPU kernels back to back, one forward pass per token, with almost no idle time. Running OLMoE-1B-7B~\cite{muennighoff2024olmoe} on \texttt{llama.cpp}, we measured $\approx 100\%$ GPU utilization on server, desktop, and embedded GPUs. Since DVFS governors raise frequency to hold utilization near a target~\cite{lin2023workload}, such load pins the clock near its peak~\cite{liu2026greenllm}; higher frequency requires higher supply voltage, so power grows superlinearly~\cite{liu2026greenllm}, raising GPU temperature~\cite{stojkovic2025tapas}.

\subsection{Temperature-Dependent IC Aging}
Electromigration~\cite{sapatnekar2019electromigration}, bias temperature instability~\cite{zhang2022bias}, and dielectric breakdown~\cite{mcpherson2023physical} accelerate with temperature~\cite{halak2019ageing}. {\color{black}We follow the industry-standard model for temperature-accelerated wear-out, the Acceleration Factor (AF)}~\cite{jesd91a2003method,ohring1998reliability}, the ratio of the Arrhenius degradation rates at a stress temperature $T_{\text{stress}}$ and a nominal operating temperature $T_{\text{normal}}$:
\begin{equation}
  AF = e^{\frac{E_a}{k}\left(\frac{1}{T_{\text{normal}}} - \frac{1}{T_{\text{stress}}}\right)},
  \label{eq:AF}
\end{equation}
where \(E_a\) is the activation energy, \(k\) is Boltzmann's constant, and temperatures are absolute; we assume \(E_a = 0.7\,\text{eV}\) for CMOS wear-out~\cite{TI-SPRABX4B,vigrass2010calculation}. For a time-varying profile with time fraction \(a_i\) and acceleration factor \(AF_i\) in interval \(i\), standard reliability analysis~\cite{jesd91a2003method} defines the \emph{effective} acceleration factor as the time-weighted sum
\begin{equation}
  AF_{\text{eff}} = \sum_i a_i AF_i.
  \label{eq:AFeff}
\end{equation}

\subsection{Proportional--Integral--Derivative Control}
A standard discrete-time PID controller~\cite{aastrom2006pid} acts on the error $e[k] = r[k] - y[k]$ between the target $r[k]$ and the measured output $y[k]$ at sampling instant $k$ through the control input
\begin{equation}
  u[k] = K_p e[k]
       + K_i \sum_{i=0}^{k} e[i] T_s
       + K_d \frac{e[k] - e[k-1]}{T_s},
  \label{eq:pid_law}
\end{equation}
where \(K_p\), \(K_i\), and \(K_d\) are the proportional, integral, and derivative gains, and \(T_s\) is the sampling period. Without requiring a thermal model, PID can track user-specified temperature or throughput targets under varying workloads.

\subsection{Mixed-Precision MoE Quantization}
FlexQuant~\cite{liu-etal-2025-flexquant} switches layer-level precision at runtime, but ranks layers by offline KL divergence and, being MoE-agnostic, treats each layer uniformly although few experts serve each token. Experts differ in usage and sensitivity~\cite{fedus2022switch,zhou2022expertchoice}, which MoE-aware schemes exploit, but they mostly fix bit-widths offline from calibration statistics~\cite{li2024quantmoebench,huang2025mcmoe,hu2025moequant}; QuantMoE~\cite{li2024quantmoebench} assigns higher precision to the experts most frequently activated on a calibration dataset, limiting generalization across inputs and domains. {\color{black}A second line sets precision at runtime: MoE-APEX~\cite{tang2026moe}, an expert-offloading system, keeps the top-routing-weight experts in high precision until their cumulative weight reaches a fixed global threshold, which leaves the high-precision count per layer uneven---some layers get many, others only one---and runs the rest in low precision to cut expert-loading latency. A layer with too few cannot carry the high-fidelity information from earlier layers, letting low-precision error dilute the representation through depth.} These limitations motivate our design principle: precision allocation should require no offline profiling, be decided per token at runtime, and balance the high-precision experts across layers.

%% file: Sections/Observation.tex
\begin{figure}[t]
	\centering
	\includegraphics[width=\linewidth]{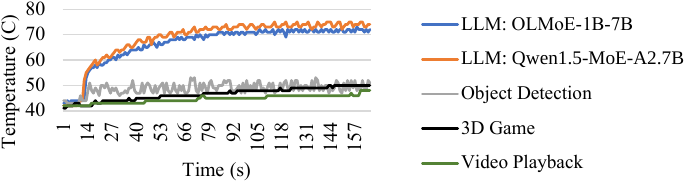}
    \vspace{-20pt}
	\caption{Temperature traces for different applications. On-device LLM inference reaches higher temperatures than other daily applications.}
	\vspace{-15pt}
\label{fig:sec3:app_temp}
\end{figure}

\section{Observation and Motivation}
\label{sec3:Motivation}
To quantify the thermal and lifetime impact of LLMs relative to other GPU applications, we collected 160 seconds of GPU temperature traces at $1\,\text{Hz}$ on a desktop RTX~4090. The workloads are two LLMs (OLMoE-1B-7B and Qwen1.5-MoE-A2.7B~\cite{teamqwen1} running the \texttt{llama-bench} benchmark of \texttt{llama.cpp}), object detection (YOLOv8~\cite{sohan2024review}), a 3D game (SuperTuxKart 1.5), and video playback (YouTube $4\text{K}\, 120\,\text{FPS}$). {\color{black}Each LLM trace spans a complete cold-start-to-plateau thermal cycle, the representative unit of an LLM interaction: starting from the idle temperature, the GPU heats up to its sustained operating plateau. We adopt this idle temperature as the normal temperature $T_{\text{normal}}$ in Eq.~\eqref{eq:AF}.} As Fig.~\ref{fig:sec3:app_temp} shows, both LLMs rapidly heat the GPU to a plateau of $70$--$75\,^{\circ}\text{C}$, whereas object detection and the 3D game stabilize near $50\,^{\circ}\text{C}$ and video playback remains the coolest.

We then map each trace to an aging metric via the effective acceleration factor in Eq.~\eqref{eq:AFeff}. The two LLMs average $66$--$68\,^{\circ}\text{C}$ with $AF_{\text{eff}} = 8.34$ and $9.78$, significantly higher than the other workloads, which average $44$--$48\,^{\circ}\text{C}$ with $AF_{\text{eff}}$ between $1.47$ and $2.06$. This demonstrates that local LLM inference accelerates aging by roughly 4--6$\times$ relative to representative GPU workloads, underscoring the urgent need for lifespan-aware thermal management in LLM inference systems. This calls for a unified runtime that adapts to dynamic thermal fluctuations while jointly preserving throughput and accuracy.

%% file: Sections/Approach.tex
\begin{figure}[t]
		\centering
		\includegraphics[width=0.72\linewidth]{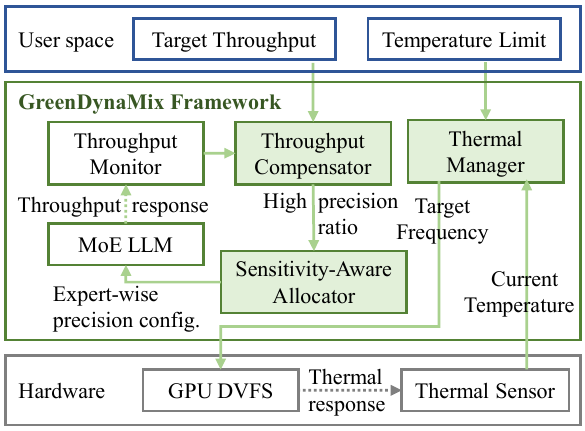}
		\vspace{-8pt}
		\caption{\textsc{GreenDynaMix} system overview.}
		\vspace{-15pt}
		\label{fig:overview}
\end{figure}

\section{GreenDynaMix}
\subsection{Design Challenge and Overview}
The central challenge in sustainable LLM inference is maintaining a low operating temperature without sacrificing throughput or accuracy. Thermal control requires dynamically adjusting the GPU frequency in response to real-time temperature fluctuations, but each frequency change also alters the available compute capacity and thus the token generation rate. Because the thermal state continues to evolve during inference, the available compute budget is inherently time-varying, requiring an equally dynamic mechanism to recover throughput. However, such compensation typically reduces per-token computation, for example, through lower arithmetic precision, which can degrade accuracy.

To address this challenge, \textsc{GreenDynaMix} combines three cooperating modules, as illustrated in Fig.~\ref{fig:overview}. Given a user-specified temperature limit and target throughput, the Thermal Manager tunes the GPU frequency, the Throughput Compensator adjusts the ratio of high- and low-precision experts to compensate for the resulting throughput loss, and the Sensitivity-Aware Allocator decides which experts receive high precision. Together, these modules maintain thermal safety while jointly preserving throughput and accuracy.

\subsection{Thermal Manager}
\label{sec:thermal_manager}
The Thermal Manager keeps the GPU within a specified thermal limit using DVFS as the control knob, since lower operating frequencies require lower supply voltages and generate less heat. Since temperature also depends on workload intensity, neighboring components, ambient conditions, and thermal inertia, a static frequency--temperature mapping is insufficient; the PID controller instead eliminates steady-state error and handles these disturbances. Let $T_{\text{limit}}$ denote the target junction temperature and $T_{\text{curr}}[k]$ the on-die sensor reading, sampled every $T_s = 1~\text{s}$. The controller uses the temperature gap as its error and interprets the PID output in Eq.~\eqref{eq:pid_law} as a \emph{frequency adjustment} (in MHz) to the previous setpoint, saturated to the allowable DVFS range $[f_{\min},\, f_{\max}]$:
\begin{equation}
  \begin{aligned}
    e[k] &= T_{\text{limit}} - T_{\text{curr}}[k],\\
    f[k] &= \operatorname{clip}\bigl(f[k-1] + u[k],\, f_{\min},\, f_{\max}\bigr).
  \end{aligned}
  \label{eq:tt-final}
\end{equation}
A temperature below the target ($e[k] > 0$) thus raises the command frequency, while $e[k] < 0$ lowers it.

\subsection{Throughput Compensator}
\label{sec:throughput_compensator}
\begin{figure}[t]
	\centering
	\includegraphics[width=0.86\linewidth]{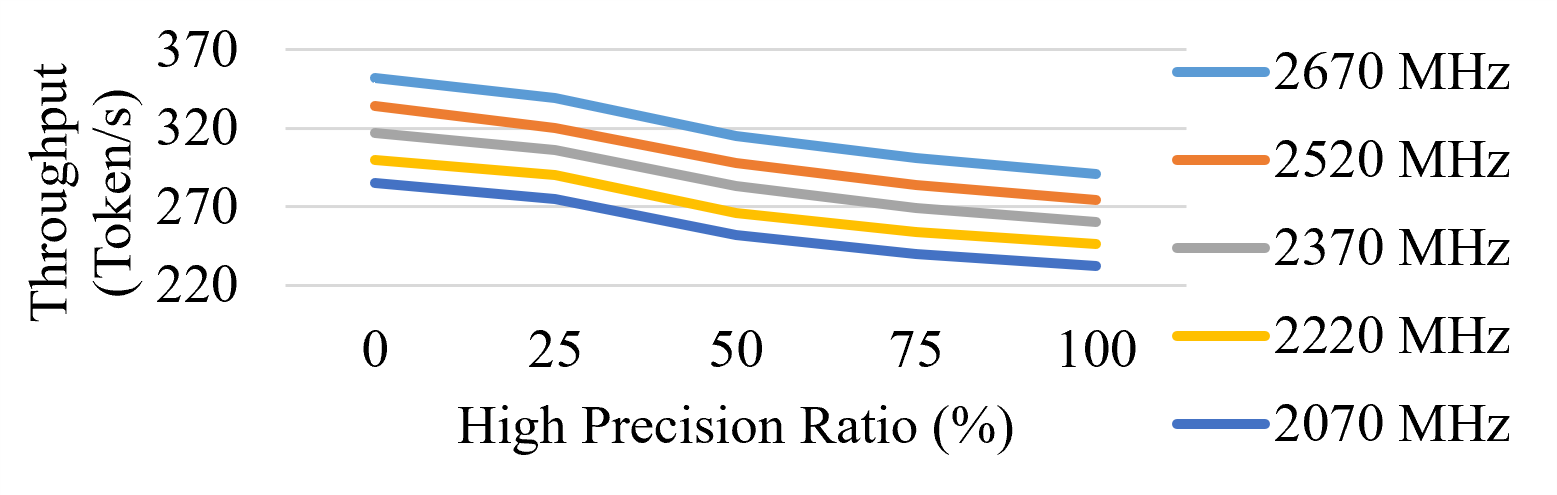}
    \vspace{-12.5pt}
	\caption{Throughput vs. high-precision ratio (0\%: all experts use low precision; 100\%: all experts use high precision).}
	\vspace{-16pt}
	\label{fig:example}
\end{figure}

Frequency reduction inevitably lowers inference throughput. To recover it under the same thermal constraint, the Throughput Compensator reduces the per-token workload through dynamic mixed-precision quantization: computing a subset of the activated experts in a lower bit-width reduces both memory traffic and arithmetic cost. As shown in Fig.~\ref{fig:example}, decreasing the high-precision ratio consistently improves token throughput across all tested frequencies, and the gain can exceed the loss from a lower frequency (e.g., the 2520\,MHz curve at 0\% lies above the 2670\,MHz curve at 100\%), so we treat this ratio as the control knob. Because throughput also depends on other system factors (e.g., GPU and memory frequency), the ratio alone does not uniquely determine it, and we again close the loop with PID control. At each control period, the measured throughput $Th_{\text{curr}}[k]$ (tokens/s) is compared against a user- or system-specified target $Th_{\text{target}}$, and the PID output updates the high-precision ratio $\rho$, i.e., the fraction of activated experts computed in high precision:
\begin{equation}
  \begin{aligned}
    e[k] &= Th_{\text{target}} - Th_{\text{curr}}[k],\\
    \rho[k] &= \operatorname{clip}\bigl(\rho[k-1] - u[k],\, 0,\, 1\bigr).
  \end{aligned}
  \label{eq:tp_clip}
\end{equation}
When the throughput is below the target ($e[k] > 0$), $\rho$ decreases so that more experts run in low precision; otherwise, $\rho$ increases to restore accuracy within the same thermal budget.

\begin{figure}[t]
	\centering
    \begin{minipage}{0.85\linewidth}
    \subfloat[]{\includegraphics[width=0.38\linewidth]{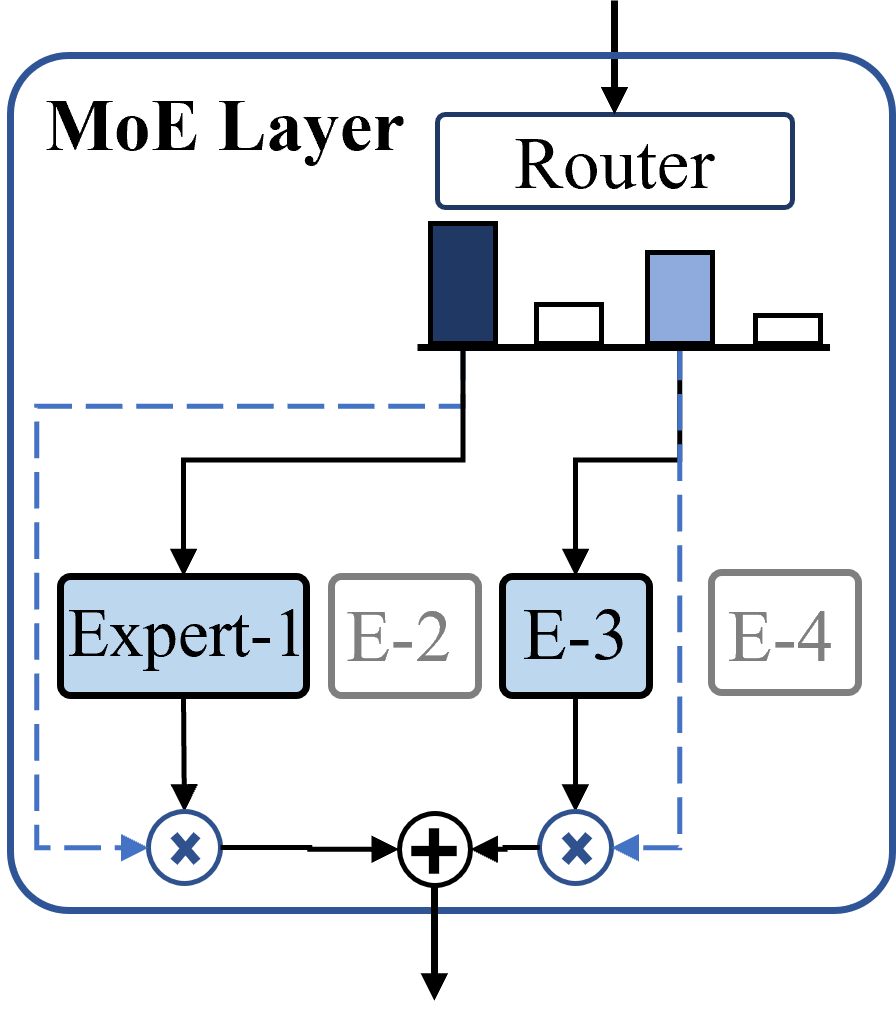}\label{fig:moe_layer_a}}\hfill
    \subfloat[]{\includegraphics[width=0.61\linewidth]{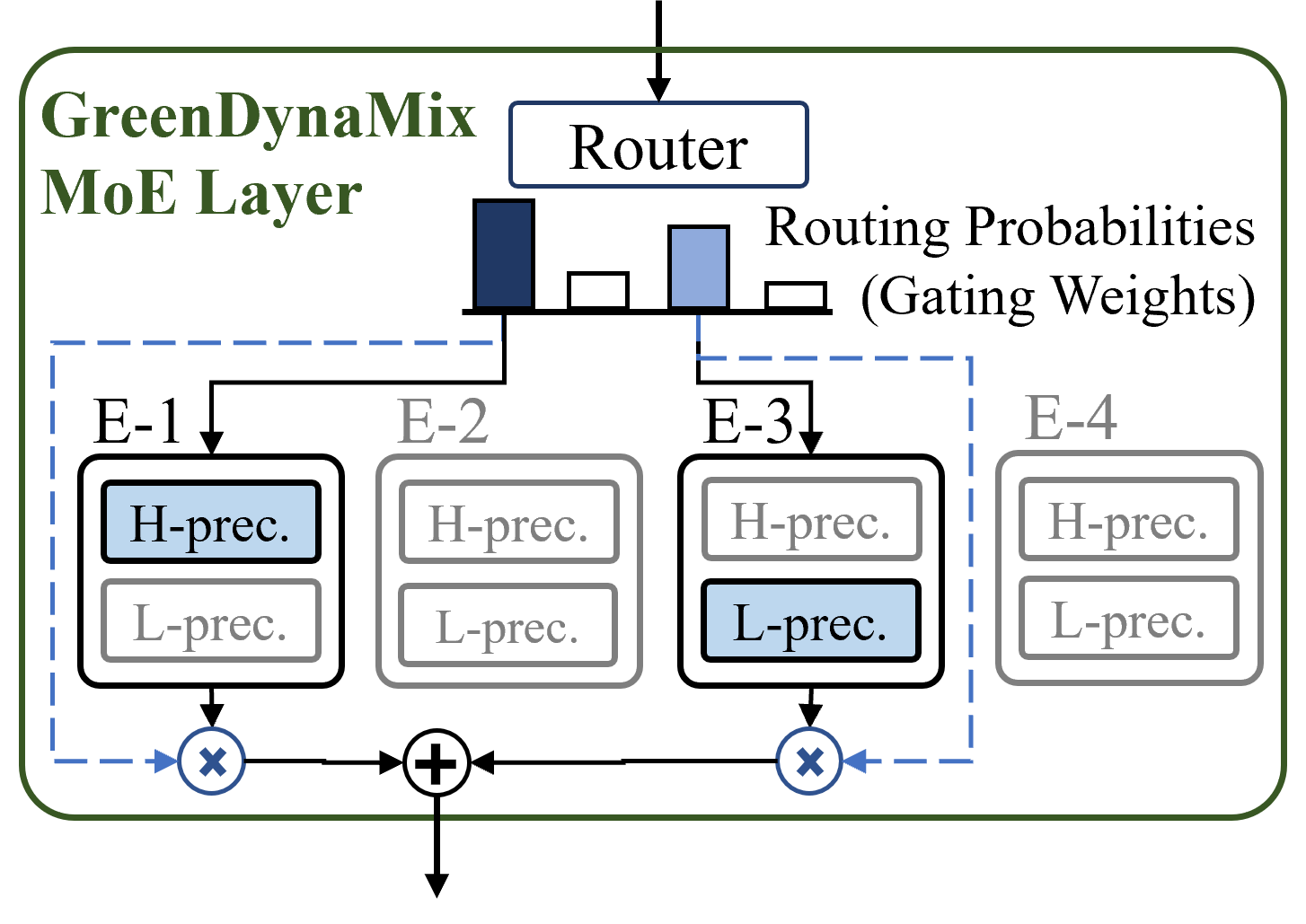}\label{fig:moe_layer_b}}
    \end{minipage}
    \vspace{-3pt}
	\caption{(a) Conventional MoE and (b) \textsc{GreenDynaMix} MoE, where each expert has high- and low-precision paths, selected by routing probabilities.}
	\vspace{-15pt}
	\label{fig:moe_layer}
\end{figure}

\begin{figure*}[t]
    \centering
    \begin{minipage}{0.85\linewidth}
    \includegraphics[width=\linewidth]{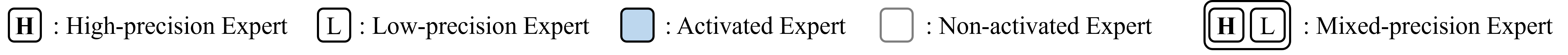}\vspace{-8pt}
    \subfloat[]{\includegraphics[width=0.3807\linewidth]{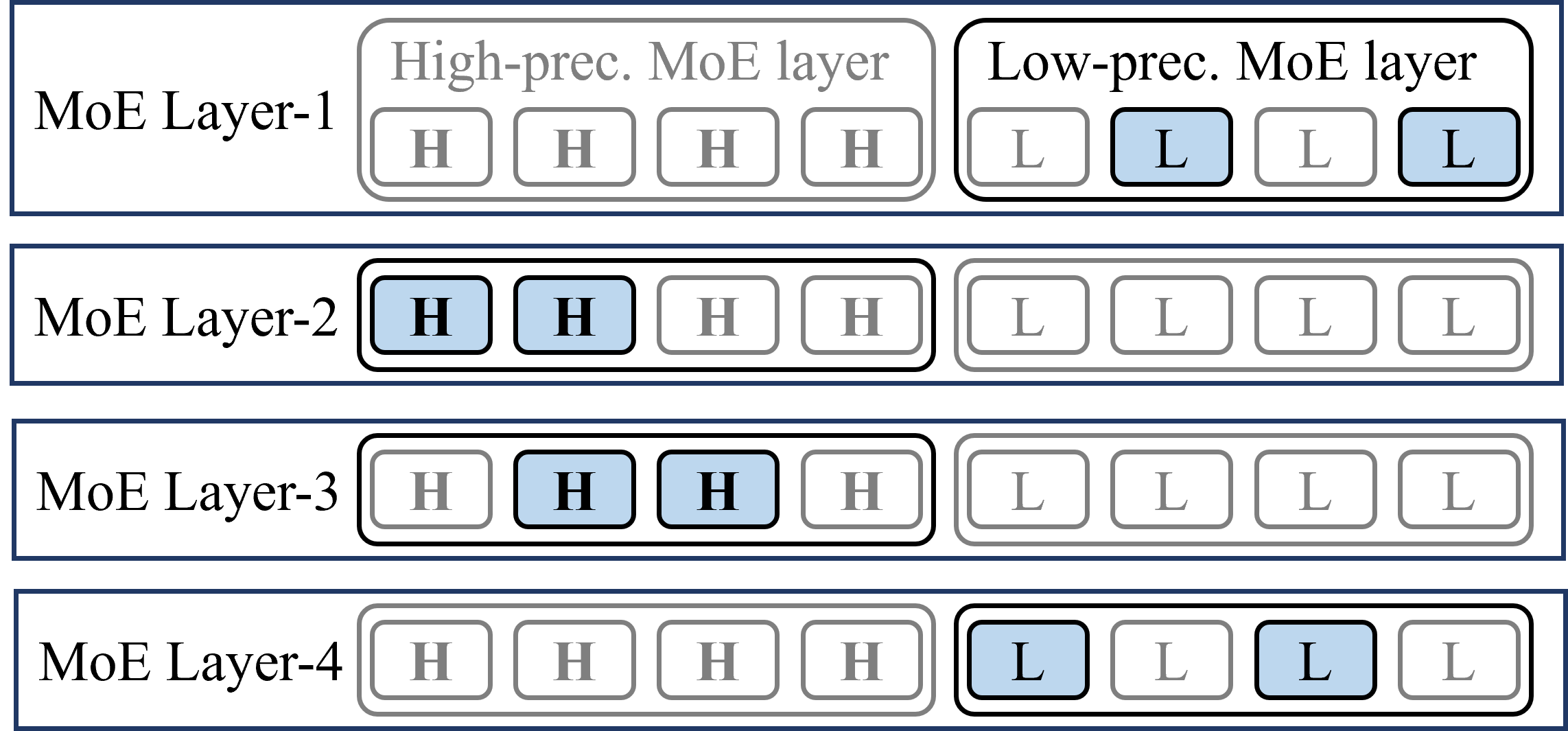}\label{fig:moe_mix_a}}\hfill
    \subfloat[]{\includegraphics[width=0.2013\linewidth]{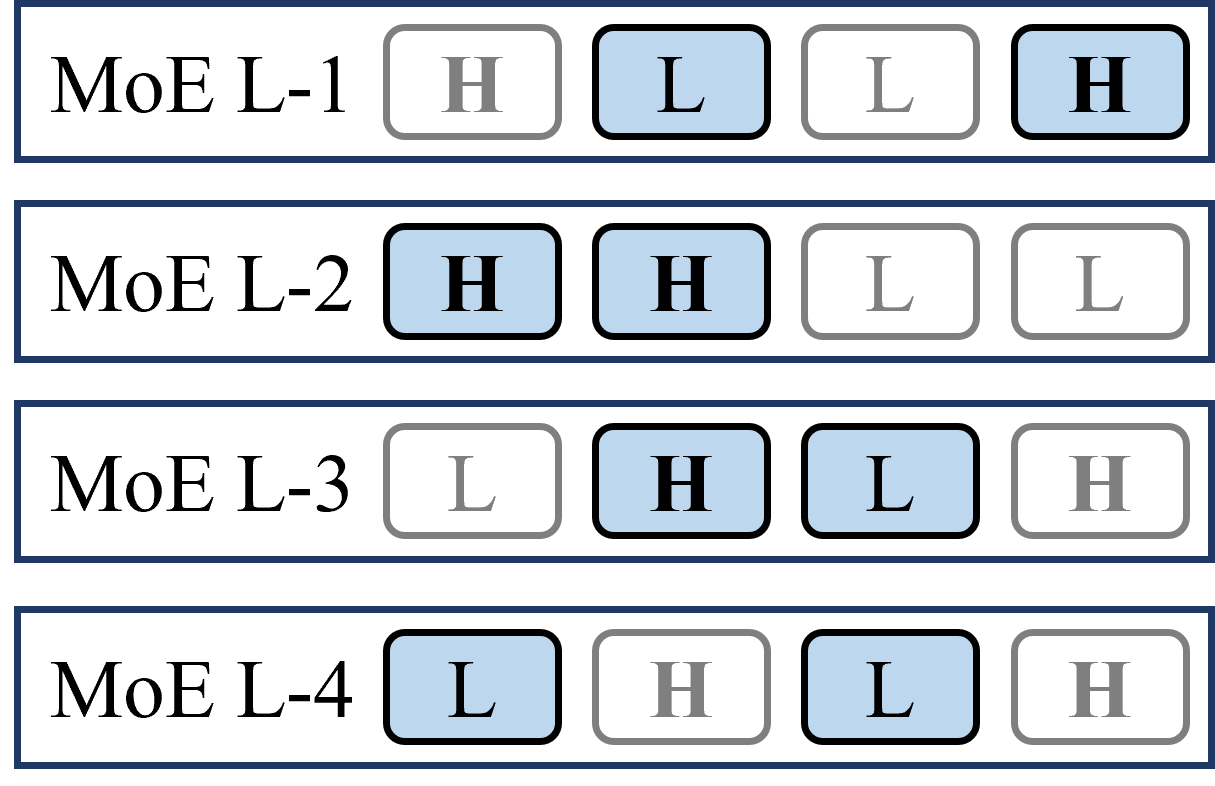}\label{fig:moe_mix_b}}\hfill
    \subfloat[]{\includegraphics[width=0.3879\linewidth]{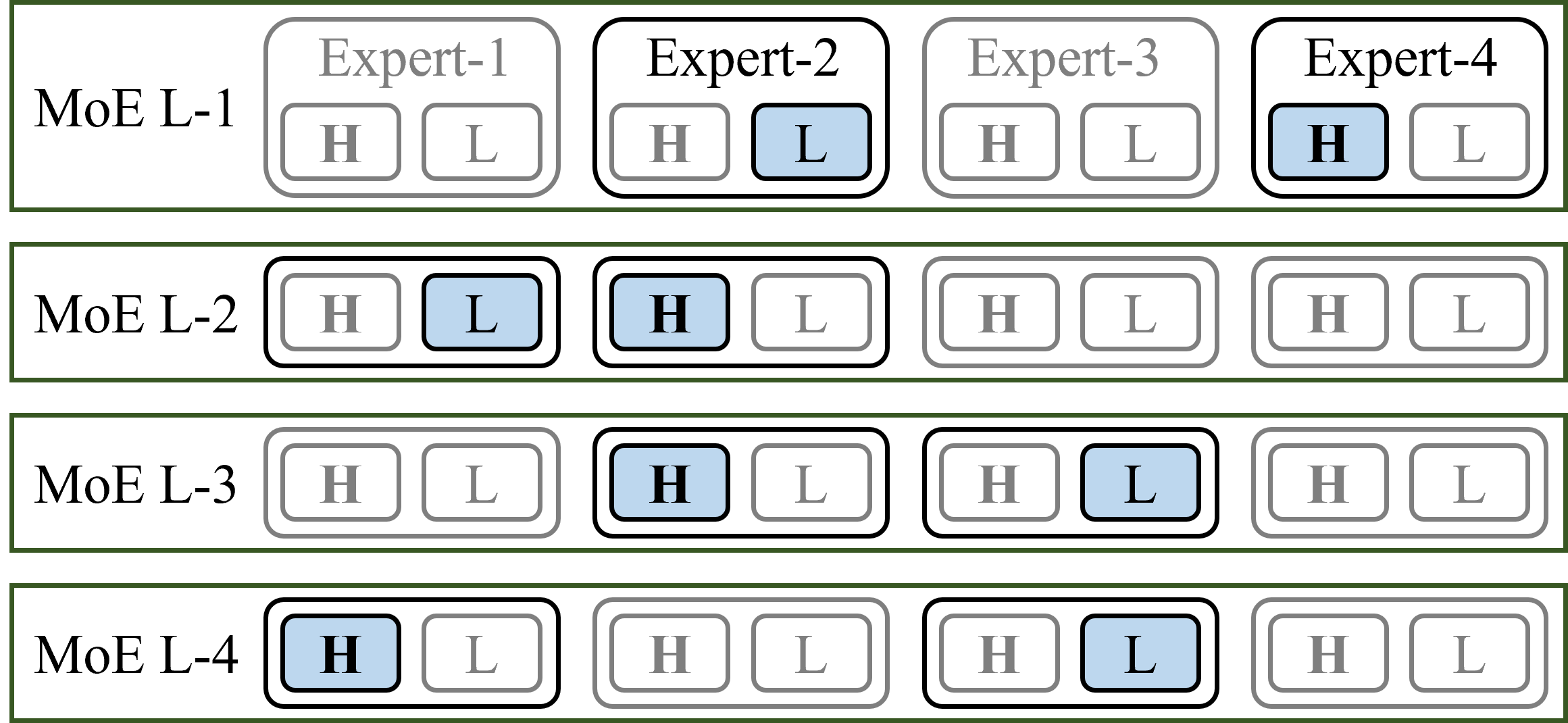}\label{fig:moe_mix_c}}
    \end{minipage}
    \vspace{-3pt}
    \caption{Comparison of high-precision budget placement strategies. (a) Dynamic layer-wise allocation. (b) Static expert-wise allocation with fixed high-precision experts. (c) \textsc{GreenDynaMix} dynamically selects high-precision paths at runtime according to router weights, where each expert supports both high- and low-precision execution.}
    \vspace{-15pt}
    \label{fig:moe_mix}
\end{figure*}

\subsection{Sensitivity-Aware Allocator}
Given $\rho$, the Sensitivity-Aware Allocator decides which activated experts run in high precision, a choice that directly impacts accuracy. As Fig.~\ref{fig:moe_layer} shows, unlike conventional MoE, where each expert has a single fixed-precision path, \textsc{GreenDynaMix} equips every expert with both high- and low-precision paths, selectable at runtime. In each MoE layer, the router produces a probability score for each expert and selects the top-$k$ experts for execution. Since experts with higher scores contribute more to the token representation, we assign high precision to the activated experts with the highest router scores. To preserve high-fidelity information across depth, we distribute the high-precision budget evenly across layers, assigning any remainder to earlier layers first so that high-precision computation propagates deeper.

Fig.~\ref{fig:moe_mix} compares this policy with prior strategies on an example with four MoE layers, four experts per layer, two activated experts per layer, and a budget of four high-precision experts. Dynamic layer-wise allocation (Fig.~\ref{fig:moe_mix_a}), as in FlexQuant~\cite{liu-etal-2025-flexquant}, sets precision for entire layers, so all activated experts in a low-precision layer are quantized, creating an information bottleneck. Static expert-wise allocation (Fig.~\ref{fig:moe_mix_b}), as in QuantMoE~\cite{li2024quantmoebench}, fixes the high-precision experts offline, so the experts activated by a given token may all fall into the low-precision set, again creating a bottleneck. {\color{black}MoE-APEX~\cite{tang2026moe} allocates precision by routing weight at runtime, but with a single global threshold, so some layers keep several high-precision experts and others only one, leaving them too narrow a high-fidelity path.} In contrast, \textsc{GreenDynaMix} (Fig.~\ref{fig:moe_mix_c}) performs runtime, router-aware allocation while distributing the budget evenly across layers, assigning high precision to the activated experts with the largest routing weights. Every layer thus preserves a high-fidelity path, allowing information to propagate through depth---avoiding the bottlenecks of the layer-wise, static, and globally thresholded schemes.

To explain the bottleneck effect, let $\mathbf{x}_{l-1}$ be the input of layer $l$, $\mathcal{A}_l$ its activated experts, $w_{l,i}$ their router weights, $f_{l,i}^{\mathrm{FP}}$ the full-precision function of expert $i$, and $\mathbf{x}_l^{\mathrm{FP}}=\sum_{i\in\mathcal{A}_l} w_{l,i} f_{l,i}^{\mathrm{FP}}(\mathbf{x}_{l-1})$ the full-precision reference output. Executing expert $i$ in high (HP) or low precision (LP) adds an error $\boldsymbol{\epsilon}_{l,i}^{\mathrm{HP}}$ or $\boldsymbol{\epsilon}_{l,i}^{\mathrm{LP}}$ to $f_{l,i}^{\mathrm{FP}}$, with $\|\boldsymbol{\epsilon}_{l,i}^{\mathrm{HP}}\| \ll \|\boldsymbol{\epsilon}_{l,i}^{\mathrm{LP}}\|$. The error of the layer output $\mathbf{x}_l$ is then
\[
\mathbf{x}_l-\mathbf{x}_l^{\mathrm{FP}}
=
\sum_{i\in\mathcal{H}_l} w_{l,i}\boldsymbol{\epsilon}_{l,i}^{\mathrm{HP}}
+
\sum_{i\in\mathcal{L}_l} w_{l,i}\boldsymbol{\epsilon}_{l,i}^{\mathrm{LP}},
\]
where $\mathcal{H}_l$ and $\mathcal{L}_l$ denote the activated experts assigned to HP and LP. If $\mathcal{H}_l=\emptyset$, the output is dominated by the larger LP error, and the hidden representation is already degraded after this layer. Such a layer is a bottleneck: all later layers compute on its degraded output. Otherwise, the layer preserves a high-fidelity component that propagates to deeper layers. Since each error is scaled by its router weight, under the same HP budget, assigning HP to the activated experts with larger router weights more effectively preserves information flow and reduces cumulative LP-induced distortion.

Formally, let there be $L$ MoE layers, and let layer $l$ activate $K_l$ experts, for a total of $N = \sum_{l=1}^{L} K_l$ activated experts. Given the ratio $\rho \in [0,1]$ from the Throughput Compensator, the total high-precision budget is $B = \lfloor \rho N \rfloor$. With $q = \lfloor B/L \rfloor$ and $r = B \bmod L$, layer $l$ receives $B_l = q + 1$ if $l \le r$ and $B_l = q$ otherwise, and runs the $H_l = \min(B_l,\, K_l)$ activated experts with the largest routing weights in high precision. In Fig.~\ref{fig:moe_mix_c}, $B = 4$ and $L = 4$ give every layer one high-precision expert ($q = 1$, $r = 0$); $B = 5$ would add a second one to the first layer.

\subsection{Implementation}
\textsc{GreenDynaMix} is platform-agnostic, requiring only (1) temperature sensing, (2) a controllable performance knob such as DVFS or an equivalent power-limit interface, (3) the token generation rate, and (4) mixed-precision execution in the inference engine; the control logic is the same on every platform. We implement it in \texttt{llama.cpp} on NVIDIA GPUs.
\subsubsection{Thermal Manager}
On server and desktop GPUs, the runtime calls \texttt{nvidia-smi} every 1\,s to read the temperature and graphics clock (\texttt{temperature.gpu}, \texttt{clocks.gr}) and to lock the clock with \texttt{-lgc} to the frequency from Sec.~\ref{sec:thermal_manager}; on Jetson platforms, \texttt{tegrastats} supplies the telemetry.
\subsubsection{Throughput Compensator}
A hook in \texttt{ggml-cuda.cu}, placed in \texttt{evaluate\_and\_\allowbreak capture\_\allowbreak cuda\_\allowbreak graph()} on the CUDA path of token generation, counts the tokens generated in each 1\,s control period; the controller in Sec.~\ref{sec:throughput_compensator} turns the throughput error into $\rho$ for the next period.
\subsubsection{Sensitivity-Aware Allocator}
Two mechanisms enable fast token- and expert-wise precision switching. (1) \textit{Dual-precision weight storage:} with the \texttt{GGUFWriter} API, we duplicate each MoE expert tensor into \texttt{\_high} and \texttt{\_low} copies, quantized separately by \texttt{llama-quantize} (e.g., \texttt{Q8\_0} and \texttt{Q4\_0}), so one \texttt{GGUF} model stores both. We register the new tensors and architecture in \texttt{llama-arch} and \texttt{llama-model}, extend \texttt{mul\_mat\_id} and \texttt{mul\_mat} in \texttt{ggml} with an optional high-precision input, and build the graph of Fig.~\ref{fig:moe_layer_b} in \texttt{llama-graph.cpp}. (2) \textit{Runtime precision dispatch:} we add a heterogeneous matrix-vector multiplication CUDA kernel in \texttt{ggml-cuda/mmvq.cu}. The budget $B_t$ for token $t$, set by the Throughput Compensator, is divided equally among the remaining layers; the experts with the highest router logits take high precision until it is spent.

%% file: Sections/Evaluation.tex
\begin{figure*}[t]
	\centering
	\includegraphics[width=\textwidth]{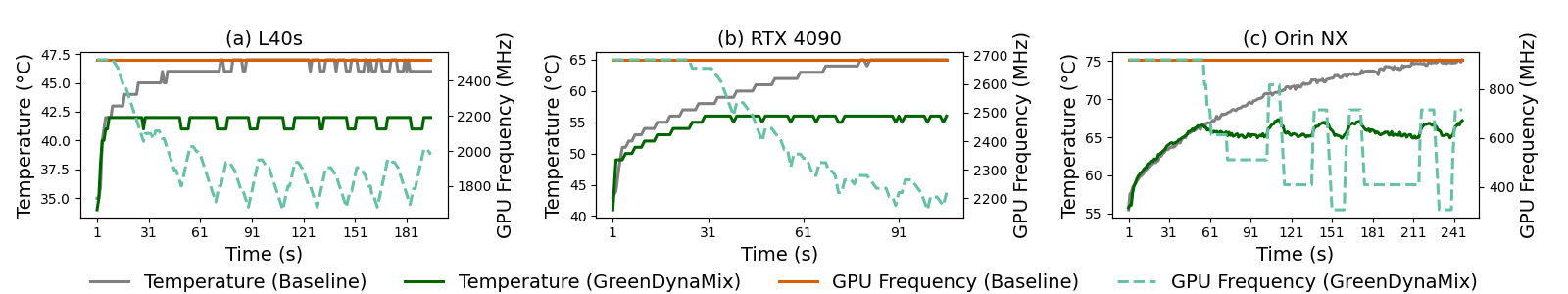}\\[2pt]
	\includegraphics[width=\textwidth]{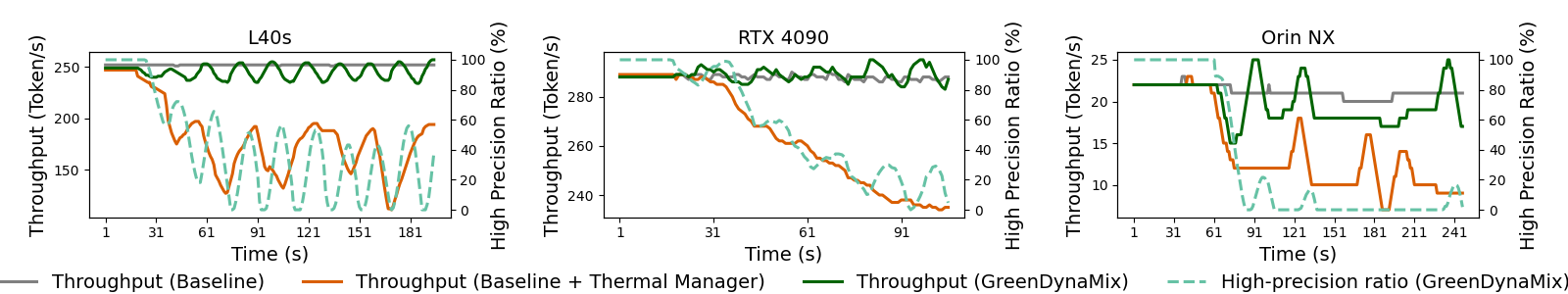}\par
    \vspace{-8pt}
    \caption{GPU temperature and frequency (top), and throughput and high-precision ratio over time (bottom), on L40S, RTX 4090, and Orin NX under the baseline and GreenDynaMix.}
    \vspace{-16pt}
	\label{fig:exp_thermal}
\end{figure*}

\begin{table}[t]
  \centering
  \caption{Thermal, aging, throughput, and accuracy metrics comparing the baseline and GreenDynaMix across devices.}
  \label{tab:overall}
  \setlength{\tabcolsep}{2pt}
  \resizebox{\columnwidth}{!}{%
  \begin{tabular}{lrccc}
    \toprule
    Metric & Config & L40S & RTX 4090 &  Orin NX \\
    \midrule
    \multirow{2}{*}{Avg. Temp (°C) $\downarrow$}
      & Baseline     & 45.8 & 60.3 & 70.0 \\
      & GreenDynaMix & 41.6 & 54.8 & 65.1 \\
    \multirow{2}{*}{Peak Temp (°C) $\downarrow$}
      & Baseline     & 47   & 65   & 75   \\
      & GreenDynaMix & 42   & 56   & 67   \\
    \midrule
    \multirow{2}{*}{Acceleration Factor $AF_{\text{eff}} \downarrow$}
      & Baseline     & 2.47 & 4.01 & 2.99 \\
      & GreenDynaMix & 1.75 & 3.01 & 2.00 \\
    \multirow{1}{*}{Aging-Rate Reduction $\uparrow$}
      & GreenDynaMix & 29\% & 25\% & 33\% \\
    \midrule

    \multirow{2}{*}{High-Prec. Ratio (\%)}
      & Baseline     & 100  & 100  & 100  \\
      & GreenDynaMix & 41   & 55   & 31   \\
    \multirow{2}{*}{Throughput (Tokens/s) $\uparrow$}
      & Baseline     & 251  & 287  & 21   \\
      & GreenDynaMix & 245  & 289  & 20   \\
    \midrule
    \multirow{2}{*}{Perplexity (PPL) $\downarrow$}
      & Baseline     & 8.8825 & 8.8825 & 9.6184 \\
      & GreenDynaMix & 9.0280 & 8.9645 & 10.5222 \\
    \bottomrule
  \end{tabular}}
\end{table}

\section{Evaluation}
\label{sec4:evaluation}
\subsection{Experimental Setup}
\subsubsection{Platforms}
We evaluate \textsc{GreenDynaMix} on three hardware tiers, running the same control algorithm without platform-specific changes: a datacenter server with an NVIDIA L40S (48\,GB), a desktop with an RTX 4090 (24\,GB), and an embedded Jetson Orin NX (16\,GB). {\color{black}For each platform, we launch an interactive \texttt{llama-cli} session (batch size~1) from the idle state and trace the GPU temperature until the baseline converges to a steady-state plateau. This cold-start-to-plateau cycle is the thermal pattern of a single prompt-triggered LLM run and spans the workload's full dynamic temperature range.}

\subsubsection{Models and Datasets}
The primary experiments use \textit{OLMoE-1B-7B}~\cite{muennighoff2024olmoe}, with $1\,\text{B}$ active and $7\,\text{B}$ total parameters, 16 MoE layers, 64 experts per layer, and top-8 expert activation; the cross-model study adds \textit{Qwen1.5-MoE-A2.7B}~\cite{teamqwen1}. High/low precision is 8-bit/4-bit on the server and desktop and 4-bit/2-bit on the embedded platform, reflecting its tighter memory and compute constraints. We report perplexity (PPL) on \textit{WikiText-2}~\cite{merity2017pointer} and \textit{Pile-10K}~\cite{Nanda2022Pile10K}, and task accuracy on \textit{HellaSwag}~\cite{zellers2019hellaswag} and \textit{WinoGrande}~\cite{sakaguchi2019winogrande}.

\subsubsection{Metrics and Baselines}
For aging, we compare $AF_{\text{eff}}$ from Eq.~\eqref{eq:AFeff} against the GPU's default-governor baseline; for throughput, we compare against a static fixed-precision baseline. {\color{black}In the budget-ratio study, we compare \textsc{GreenDynaMix} against three high-precision-budget allocation policies at matched budgets: FlexQuant~\cite{liu-etal-2025-flexquant} (dynamic but MoE-agnostic, layer-wise), QuantMoE~\cite{li2024quantmoebench} (MoE-aware but offline, statically profiled, expert-wise), and the router-guided allocation of MoE-APEX~\cite{tang2026moe} (runtime, expert-wise).}

\setlength{\tabcolsep}{2pt}
\begin{table*}[t]
\centering
\caption{Cross-model ablation study on RTX 4090 using OLMoE and Qwen.}
\label{tab:cross_model_ablation}
\resizebox{\textwidth}{!}{
\begin{tabular}{ccc|ccccc|ccccc}
\toprule
\multirow{2}{*}[-0.5ex]{\shortstack{Thermal\\Manager}} &
\multirow{2}{*}[-0.5ex]{\shortstack{Throughput\\Compensator}} &
\multirow{2}{*}[-0.5ex]{\shortstack{Sensitivity-Aware\\Allocator}} &
\multicolumn{5}{c|}{OLMoE-1B-7B} &
\multicolumn{5}{c}{Qwen1.5-MoE-A2.7B} \\
\cmidrule(lr){4-8} \cmidrule(l){9-13}
& & &
Peak Temp$\downarrow$ &
$AF_{\mathrm{eff}}\downarrow$ &
Aging-Rate Red.$\uparrow$ &
Throughput$\uparrow$ &
HellaSwag$\uparrow$ &
Peak Temp$\downarrow$ &
$AF_{\mathrm{eff}}\downarrow$ &
Aging-Rate Red.$\uparrow$ &
Throughput$\uparrow$ &
HellaSwag$\uparrow$ \\
\midrule
$\times$      & $\times$      & $\times$      & 65 & 4.01 & - & 287 & 80.24\% & 65 & 3.79 & - & 239 & 75.69\% \\
$\checkmark$  & $\times$      & $\times$      & 56 & 3.01 & 25\% & 257 & 80.24\% & 56 & 2.79 & 26\% & 226 & 75.69\% \\
$\checkmark$  & $\checkmark$  & $\times$      & 56 & 3.01 & 25\% & 289 & 80.05\% & 56 & 2.79 & 26\% & 240 & 75.14\% \\
$\checkmark$  & $\checkmark$  & $\checkmark$  & 56 & 3.01 & 25\% & 289 & 80.13\% & 56 & 2.79 & 26\% & 240 & 75.49\% \\
\bottomrule
\end{tabular}
}
\vspace{-12pt}
\end{table*}
\subsection{Evaluation Results}
\subsubsection{Temperature \& Aging Rate}
Fig.~\ref{fig:exp_thermal} (top) compares GPU temperature traces for the default governor (baseline) and \textsc{GreenDynaMix}. The baseline runs at maximum frequency and settles at a high, stable temperature plateau, whereas the Thermal Manager modulates the GPU frequency to track a target temperature set at the midpoint between the idle state and the baseline steady state. Using these traces, Table~\ref{tab:overall} shows that \textsc{GreenDynaMix} consistently reduces the Arrhenius acceleration factor: from $2.47$ to $1.75$ (L40S), $4.01$ to $3.01$ (RTX 4090), and $2.99$ to $2.00$ (Orin NX), corresponding to aging-rate reductions of $29\%$, $25\%$, and $33\%$ during inference. The temperature falls because frequency control lowers the average GPU power; for example, we measured a drop from $228.7$ to $167.9$~watts ($27\%$) on the L40S and from $265.5$ to $220.8$~watts ($17\%$) on the RTX~4090.

\subsubsection{Throughput}
Fig.~\ref{fig:exp_thermal} (bottom) compares the throughput of the baseline, the baseline with only the Thermal Manager enabled, and \textsc{GreenDynaMix}. We set $Th_{\text{target}}$ to the baseline's average throughput on each platform. Enabling only the Thermal Manager reduces temperature, but DVFS causes throughput to steadily drop. \textsc{GreenDynaMix} undergoes the same frequency reduction, but its Throughput Compensator recovers the lost throughput by mixing high- and low-precision experts, keeping the average rate comparable to the baseline on all platforms (Table~\ref{tab:overall}).

\setlength{\tabcolsep}{2pt}
\begin{table}[t]
\centering
\caption{Quality comparison under different high-precision activation ratios.}
\label{tab:perplexity}
\resizebox{\columnwidth}{!}{%
\begin{tabular}{l l c c c c c}
\toprule
\textbf{Benchmark} & \textbf{Method} & \textbf{0\%} & \textbf{25\%} & \textbf{50\%} & \textbf{75\%} & \textbf{100\%} \\
\midrule
\multirow{4}{*}{WikiText-2 (PPL) $\downarrow$}
& FlexQuant$^{\dagger}$   & 9.6184 & 9.5016 & 9.2394 & 9.0630 & 8.8825 \\
& QuantMoE$^{\dagger}$    & 9.6184 & 9.4458 & 9.2778 & 9.0514 & 8.8825 \\
& MoE-APEX                & 9.6184 & 9.1349 & 8.9779 & 8.9081 & 8.8825 \\
& GreenDynaMix            & 9.6184 & \textbf{{\underline{9.1294}}} & \textbf{{\underline{8.9692}}} & \textbf{{\underline{8.9066}}} & 8.8825 \\
\midrule
\multirow{4}{*}{Pile-10K (PPL) $\downarrow$}
& FlexQuant$^{\dagger}$   & 24.2880 & 23.9426 & 23.5729 & 23.0902 & 22.5846 \\
& QuantMoE$^{\dagger}$    & 24.2880 & 23.9879 & 23.5807 & 23.1669 & 22.5846 \\
& MoE-APEX                & 24.2880 & 23.1799 & 22.8445 & 22.6473 & 22.5846 \\
& GreenDynaMix            & 24.2880 & \textbf{{\underline{23.0960}}} & \textbf{{\underline{22.7718}}} & \textbf{{\underline{22.6368}}} & 22.5846 \\
\midrule
\multirow{4}{*}{WinoGrande (\%) $\uparrow$}
& FlexQuant$^{\dagger}$   & 66.38 & 66.39 & 67.72 & 68.51 & 68.75 \\
& QuantMoE$^{\dagger}$    & 66.38 & 66.46 & 68.03 & \textbf{{\underline{68.67}}} & 68.75 \\
& MoE-APEX                & 66.38 & 67.64 & 67.56 & 68.51 & 68.75 \\
& GreenDynaMix            & 66.38 & \textbf{{\underline{68.90}}} & \textbf{{\underline{68.35}}} & 68.19 & 68.75 \\
\bottomrule
\end{tabular}%
}
\begin{flushleft}
\footnotesize $^{\dagger}$Requires offline profiling.
\end{flushleft}
\vspace{-12pt}
\end{table}

\subsubsection{Accuracy}
Table~\ref{tab:overall} reports the high-precision ratio and perplexity averaged over each run. On the server and desktop (8-bit/4-bit), \textsc{GreenDynaMix} reduces the high-precision ratio to 41\% and 55\% while incurring only 1.6\% and 0.9\% perplexity degradation, respectively. On the embedded Orin NX, the tighter budget forces a more aggressive configuration (4-bit/2-bit) and a 69\% reduction in the high-precision ratio, yielding a larger 9.4\% degradation. Table~\ref{tab:perplexity} compares the allocation policies at the intermediate budgets, since all methods coincide at the $0\%$ and $100\%$ endpoints by construction. \textsc{GreenDynaMix} attains the lowest perplexity on both WikiText-2 and Pile-10K across all budgets ($25\%$--$75\%$). On WinoGrande, it leads at $25\%$ and $50\%$, already matching full-precision accuracy with only a 25\% budget; at higher budgets all methods converge to within a fraction of a point of full precision. Router-aware runtime allocation is thus more effective than both dynamic but MoE-agnostic and MoE-aware but static allocation. It also matches or lowers the perplexity of MoE-APEX---the only other runtime, router-weight--based baseline---at every budget, owing to its per-layer budget distribution rather than a single global threshold.

\subsubsection{Cross-Model Validation}
Table~\ref{tab:cross_model_ablation} reports a cross-model ablation on the RTX 4090; \textit{OLMoE-1B-7B} and \textit{Qwen1.5-MoE-A2.7B} exhibit the same trend. The Thermal Manager alone reduces peak temperature from $65\,^{\circ}\mathrm{C}$ to $56\,^{\circ}\mathrm{C}$ for both models and lowers $AF_{\mathrm{eff}}$, yielding aging-rate reductions of $25\%$ for OLMoE and $26\%$ for Qwen, but at the cost of throughput. The Throughput Compensator restores throughput to the baseline level or slightly above it while preserving the same thermal and aging benefits. Finally, the Sensitivity-Aware Allocator improves accuracy under the same thermal constraint and throughput target, increasing HellaSwag from $80.05\%$ to $80.13\%$ on OLMoE and from $75.14\%$ to $75.49\%$ on Qwen, showing that \textsc{GreenDynaMix} is not model-specific.

\subsection{System Overhead Analysis}
\subsubsection{Runtime Latency}
Timed with C \texttt{clock()} on the RTX 4090, the Thermal Manager, Throughput Compensator, and Sensitivity-Aware Allocator incur about $450\,\mu\text{s}$, $2\,\mu\text{s}$, and $32\,\mu\text{s}$ per iteration, respectively; the Thermal Manager dominates because it reads the thermal sensor, queries the GPU frequency, and applies a new setting. A full iteration takes about $484\,\mu\text{s}$, only $0.048\%$ of the 1\,s control period.
\subsubsection{Memory Footprint}
To reduce runtime dispatch to a lightweight selection, \textsc{GreenDynaMix} preloads both precision variants of the expert weights. For OLMoE, the original all-expert 8-bit model occupies 7,973 MB, whereas \textsc{GreenDynaMix} with preloaded 8-bit/4-bit expert weights occupies 11,333 MB, a 42.1\% increase; for Qwen, the footprint grows from 16,489 MB to 23,087 MB, a 40.0\% increase. In exchange, runtime precision switching has near-zero latency. This cost is what restricts the Orin NX, whose 16\,GB is shared with the operating system, to 4-bit/2-bit, and its 2-bit experts account for the larger perplexity loss in Table~\ref{tab:overall}. The duplication is not inherent to the design: a memory-constrained device could keep only the high-precision representation and derive low-precision variants on demand, following any-precision or Matryoshka quantization~\cite{park2024anyprecision,nair2025matryoshka}, or preload only the frequently reused ones. On embedded devices, however, limited compute may make this derivation a challenge worth addressing in future work, so that devices such as the Orin NX can hold a single 8-bit model and switch between 8-bit and 4-bit.

%% file: Sections/Conclusion.tex
\section{Conclusion}
\label{sec5:Conclusion}
We presented \textsc{GreenDynaMix}, a thermal-aware dynamic mixed-precision runtime for MoE LLMs that jointly manages temperature, throughput, and accuracy through a DVFS-based Thermal Manager, a Throughput Compensator that adjusts the high-/low-precision ratio, and a Sensitivity-Aware Allocator that leverages MoE routing weights. Across server, desktop, and embedded GPUs, \textsc{GreenDynaMix} reduces the temperature-induced aging rate during LLM inference by 25--33\%, while maintaining throughput close to the baseline at a perplexity cost of 0.9--1.6\% with 8-bit/4-bit mixed precision, rising to 9.4\% where memory limits the device to 4-bit/2-bit. It generalizes across MoE architectures and requires no offline profiling, retraining, or fine-tuning.